\documentclass{article}
\usepackage{authblk}
\usepackage{geometry}
\usepackage{microtype}
\usepackage{graphicx}
\usepackage{subfigure}
\usepackage{booktabs} %
\usepackage[autostyle, english = american]{csquotes}
\MakeOuterQuote{"}

\usepackage{hyperref}

\usepackage{xcolor} 

\usepackage{amsmath, amsthm, amssymb}
\usepackage{float}
\usepackage{siunitx}

\usepackage{hyperref}       %
\usepackage{url}            %
\usepackage{booktabs}       %
\usepackage{amsfonts}       %
\usepackage{nicefrac}       %
\usepackage{microtype}      %
\usepackage{xcolor}         %
\usepackage{graphicx}

\usepackage[textsize=tiny]{todonotes}

\usepackage{natbib}

\title{ExDAG: An MIQP Formulation and Algorithm for Learning DAGs}
\author[1]{Pavel Rytíř}
\author[1]{Aleš Wodecki}
\author[1]{Jakub Mareček}

\affil[1]{Czech Technical University in Prague}

\begin{document}
\maketitle

\begin{abstract}
There has been a growing interest in causal learning in recent years. Commonly used representations of causal structures, including Bayesian networks and structural equation models (SEM), take the form of directed acyclic graphs (DAGs). 
We provide a novel mixed-integer quadratic programming formulation and an associated algorithm that identifies DAGs with a low structural Hamming distance between the identified DAG and the ground truth, under identifiability assumptions. 
The eventual exact learning is guaranteed by the global convergence of the branch-and-bound-and-cut algorithm, which is utilized.
In addition to this, integer programming techniques give us access to the dual bound, which allows for a real time assessment of the quality of solution.
Previously, integer programming techniques have been shown to lead to limited scaling in the case of DAG identification due to the super exponential number of constraints, which prevent the formation of cycles.
The algorithm proposed circumvents this by selectively generating only the violated constraints using the so-called "lazy" constraints methodology. Our empirical results show that ExDAG outperforms state-of-the-art solvers in terms of structural Hamming distance and $F_1$ score when considering Gaussian noise on medium-sized graphs.
\end{abstract}

\section{Introduction}
Learning of causal representations and causal inference has received significant attention recently \citep[e.g.]{peters2017elements,9363924,ahuja2023interventional}.
With the aim of tackling a variety of challenges in a wide range of applications, many models and methodologies \citep[e.g.]{pearl_2009,park2023measure,buchholz2024learning,lorch2024causal, DAGswithNoCurl, zhang2024analytic, pmlr-v139-chen21c} have been introduced. 
In machine learning literature, probabilistic graphical models \citep{Koller09}, in general, and Bayesian networks, in particular, are often used.
In statistics and biomedical applications,
structural equation models \citep{yuan200610,duncan2014introduction} and additive noise models \citep{peters2017elements} are widely used.
All of the aforementioned can be seen as learning of (edge-weighted) directed acyclic graphs (DAGs).

The learning of DAGs, where the vertices correspond to the random variables and the oriented edges represent their dependencies,
underlies the learning of both Bayesian networks and structural equation models, where algebraic manipulations can be interpreted as interventions in the causal system
\citep{bottou2013counterfactual}. The identification of such a structure is usually mediated by a score function whose minimization identifies a class of graphs. Alternatively, selective model averaging may be used \citep{mad94}. 

In the present article, we focus on the learning of a DAG using a polynomial score function under the assumption of identifiability, which is given by persistent excitation \citep[Section 2]{willems2005note}, or equivalently, a rank condition on the Henkel matrix \citep[Theorem 1]{willems2005note}. Depending on the construction of the score function \citep{heckerman2022tutorial}, the score-optimal DAG maximizes the likelihood for Gaussian and non-Gaussian noise. 

The paper is structured as follows. Section \ref{sec_related_work} contains a brief overview of the state-of-the-art, which is most relevant to the presented work along with the motivation that led to the development of ExDAG. The following two sections (Sections \ref{sec_problem_formulation} and \ref{sec_identification_MIQP}) detail the model used for identification, identifiability assumptions, and a description of the algorithms, which follows from the formulation of the problem as a mixed integer quadratic optimization. Lastly, Section \ref{sec_numerics} contains detailed benchmarks that contextualize the performance of ExDAG with respect to the state-of-the-art.

We note that various extensions of these models abound. In particular, the NeurIPS 2023 competition organized by the gCastle team \citep{zhang2021gcastle} focused on learning the causal structure from event sequences with prior knowledge. Without prior knowledge, we improve the results of NOTEARS on the only public dataset. A direct comparison with the winners \citep{XHS,FAKEiKUN,ustc-aig} is nontrivial, as the other datasets remain private.

\subsection{Main Contributions}
Our contributions within the learning of a DAG, such as in the learning of a
Bayesian network, comprise the following. 
\begin{itemize}
    \item We consider the identification of directed acyclic graphs (DAGs) that leads to a mixed-integer quadratic problem, which is a natural choice since the form of the cost function guarantees that the global solution of the problem is a maximum likelihood estimator (see Section \ref{sec_problem_formulation}).
    \item We propose ExDAG, a branch-and-bound-and-cut algorithm for solving the formulation, which avoids the use of exponentially many constraints at the root node and avoids exponential-time preprocessing steps, while making use of mixed-integer quadratic programming techniques, which lead to a guarantee of global convergence.
    \item We perform a comparison of ExDAG with state-of-the-art solvers of \cite{and24,zheng2018dags,wax24}, which have appeared in major venues. The comparison shows favorable structural Hamming distance and $F_1$ score figures for problems with up to 25 variables.
\end{itemize}

\section{Related Work}\label{sec_related_work}
Focusing on the identification of Bayesian networks representing causal inference, we may delineate between two major strategies that lead to identification. The first of these focuses on global optimality, which, due to the presence of cycle-exclusion constraints, leads to an exponential running time \citep[e.g.]{cussens2012bayesian,bartlett2017integer, cussens2017polyhedral,studeny2021dual,kitson23}. This naturally leads to scaling issues, which are typically overcome by imposing additional assumptions such as a maximal degree or certain structural properties of the graph (e.g., existence of a decomposition from structural graph theory, cf. \cite{hlinveny2008width}). This greatly limits the applicability of global methods in practice as many problems of interest do not conform to these assumptions.

Recent locally convergent methods \citep{zheng2018dags,and24,wax24} address the aforementioned shortcoming by formulating the problem as a continuous optimization. Cycle exclusion is then enforced by means of a continuous function of the adjacency matrix, which allows scaling to hundreds of vertices. This scaling, however, is bought at a price as the convergence to global optimum is no longer guaranteed (due to non-convex constraints) and results may vary based on problem set. For certain problem sets, these solvers enjoy reasonable precision, for others not so much. The aforementioned sensitivity to instance selection is further documented in Section~\ref{sec_numerics}.

In this paper, we present a novel method which comes with global optimality guarantees while still allowing for scaling. The basis for the algorithmic solution is the structural equation model (SEM) described in the following section.

\section{Problem Formulation and Identification Results}\label{sec_problem_formulation}
In general, the problem of score-based Bayesian network learning can be seen the identification of a structural equation model (SEM), see \citep{hoover03, killian11}. Assuming linearity, no backwards in time dependence between the $d$ random variables (autoregressive order zero), the model is defined as follows.

Let $d$ be the number of variables. Let $n$ be the number of samples. Let $W\in\mathbb{R}^{d\times d}$ be the weighted adjacency matrix of the underlying directed acyclic graph (DAG) that describes the dependencies of the variables of the Bayesian network. Then the structural equation~\cite{pearl_2009} equals
\begin{equation}\label{eq:sem}
X = XW + Z,
\end{equation}
where $X=\left(X_1,\dots,X_d\right)$ is the vector of real random variables and $Z \in \mathbb{R}^d$ is an additive noise vector.

The identification (learning) problem is formulated as follows.
Let $X_\text{data}$ be a real matrix containing $n$ data samples for each of $d$ variables, $X\in \mathbb{R}^{n\times d}$. Then the goal of the learning probem is to find acyclic $W\in\mathbb{R}^{d\times d}$ that minimizes the Frobenius norm of $Z_\text{data}$ while the following equation is satisfied.
\begin{equation}\label{eq:sem_data}
X_\text{data} = X_\text{data}W + Z_\text{data},
\end{equation}
where  $Z_\text{data} \in \mathbb{R}^{n\times d}$ is an additive noise matrix.

The above can be rewriten as the minimization of the following cost function.
\begin{equation}
J\left(W\right)=\left\Vert X_\text{data}-X_\text{data}W\right\Vert _{F}^{2}+\lambda\left\Vert W\right\Vert.
\end{equation}
where $\lambda > 0$ is a (small) regularization coefficient, $\left\Vert \cdot\right\Vert $ denotes an arbitrary norm, which is usually chosen to be the $l1$-norm and $\left\Vert \cdot\right\Vert _{F}$ denotes denotes the Frobenius norm.

More concisely, the problem of score based DAG learning is defined as
\begin{align}\label{eq_DAG_base_form}
\begin{split}
    \underset{W\in\mathbb{R}^{d\times d}}{\min} J\left(W\right), \\
W \text{ is acyclic}.     
\end{split}  
\end{align}

Under a variety of identifiability assumptions, it has been shown that the solution of \eqref{eq_DAG_base_form} recovers a DAG from a given equivalence class with high probability under Gaussian \citep{geer12, aragam2017learning} and non-Gaussian noise vectors \citep{Shimizu2006ALN,loh13}. We refer to \cite{willems2005note,ahuja2023interventional} for a further discussion of identification.

\section{Identification as a Mixed Integer Quadratic Problem}\label{sec_identification_MIQP}
Let us present an MIQP formulation for the DAG learning problem. The construction is such that the cycle-exclusion constraints can be added during the runtime using an MIQP callback. This is crucial when scaling to larger instances. Another key feature is that such a callback makes use of a simple separation routine that has only quadratic complexity in the number of graph vertices in the worst case.

First, we define MIQP variables. Let $d$ be the number of random variables. The binary variables $e_{ij}$ encode a directed graph $G=(V,E)$, where $V=\{1,\dots,d\}$ and $E\subseteq V\times V$. Their values are interpreted as follows:
\begin{equation}\label{eq_decision_integer}
    e_{ij}=\begin{cases}
        0, & \text{if $(i,j)\notin E$,}\\
        1, & \text{if $(i,j)\in E$,}
    \end{cases}
\end{equation}
where $i=1,\dots,d$ and $j=1,\dots,d$.

The weights $w: E\mapsto\mathbb{R}$ of the graph $G$ are represented by the continuous variables $w_{ij}$ for $i=1,\dots,d$ and $j=1,\dots,d$.

Next, we define MIQP constraints that enforce the acyclicity of $G$. Let $C$ be the set of all possible cycles of a graph $G$ on $d$ vertices. 
Let $c\in C$ be a cycle. Then $E(c)=\left\{(i_{1},i_{2}), ({i_{2},i_{3}}),\ldots,({i_{k-1},i_{1})}\right\}$ equals the set of edges contained in $c$.

For each cycle $c\in C$, we define the following constraint, that prohibits cycle $c$ to appear in $G$:
\begin{equation}\label{eq_constraints_mi}
\sum_{(i, j) \in E(c)}e_{ij}\leq \lvert E(c)\rvert -1.
\end{equation}

Note that there are exponentially many cycles in a complete graph on $d$ vertices. Therefore, the number of constraints above~\eqref{eq_constraints_mi} is also exponential. We discuss how to reduce this number in Section~\ref{sec_quadratic_problem_callback}.

Next, we add the following constraints that force $w_{ij}$ to be zero whenever $e_{ij}$ is zero:
\begin{equation}\label{eq:w}
\begin{split}
    w_{ij}&\leq ce_{ij}, \\
    w_{ij}&\geq-ce_{ij},
\end{split}
\end{equation}
for all $i,j\in\left\{ 1,\ldots,d\right\}$, where $c>0$ denotes a constant that corresponds to the largest weight magnitude allowed. In our experiments, we set $c$ equals 100, which was sufficiently large, given our knowledge about the dataset. In the case that maximum weight is not known, one can perform experiments with various $c$ and check if the solution changes. 

Finally, we define the MIQP objective function. Let $n$ be the number of data samples. Then the objective equals:

\begin{equation}\label{eq_cost_function_with_p}
J(W)=\sum_{l=1}^{n}\sum_{i=1}^{d}\left(X_{li}-\sum_{j=1}^dX_{lj}w_{ji}\right)^{2}+\lambda\sum_{i=1}^d\sum_{j=1}^de_{ij}.
\end{equation}
We dropped the index "data" from $X_\text{data}$ in order to simplify the equation.
The regularization constant $\lambda > 0$ in \eqref{eq_cost_function_with_p} is discussed in Section \ref{sec_numerics}.

\subsection{The Branch-and-Bound-and-Cut Algorithm}\label{sec_quadratic_problem_callback}

A key contribution of ours is a branch-and-bound-and-cut algorithm to solve the formulation above.  
We utilize the usual branch-and-bound algorithm \citep[e.g.]{achterberg2007}, but implement cycle exclusion \eqref{eq_constraints_mi} using the so-called ``lazy'' constraints. Lazy constraints are only checked when an integer-feasible solution candidate has been identified. When a lazy constraint is violated, it is included across all nodes of the branch-and-bound tree.

In summary, at the root node, we utilize only $O(|E|)$ constraints \eqref{eq:w}. Subsequently, the solver introduces cycle-exclusion constraints \eqref{eq_constraints_mi}. The number of cycle exclusion constraints is possibly super-exponential, but numerical practice has shown that the number of cycle constraints needed is far lower than the "full" super-exponential amount.
This feature is supported by many open source and industrial solvers like CPLEX, GUROBI or SCIP.

\begin{figure*}[h]
    \centering
    \includegraphics[width=0.9\linewidth]{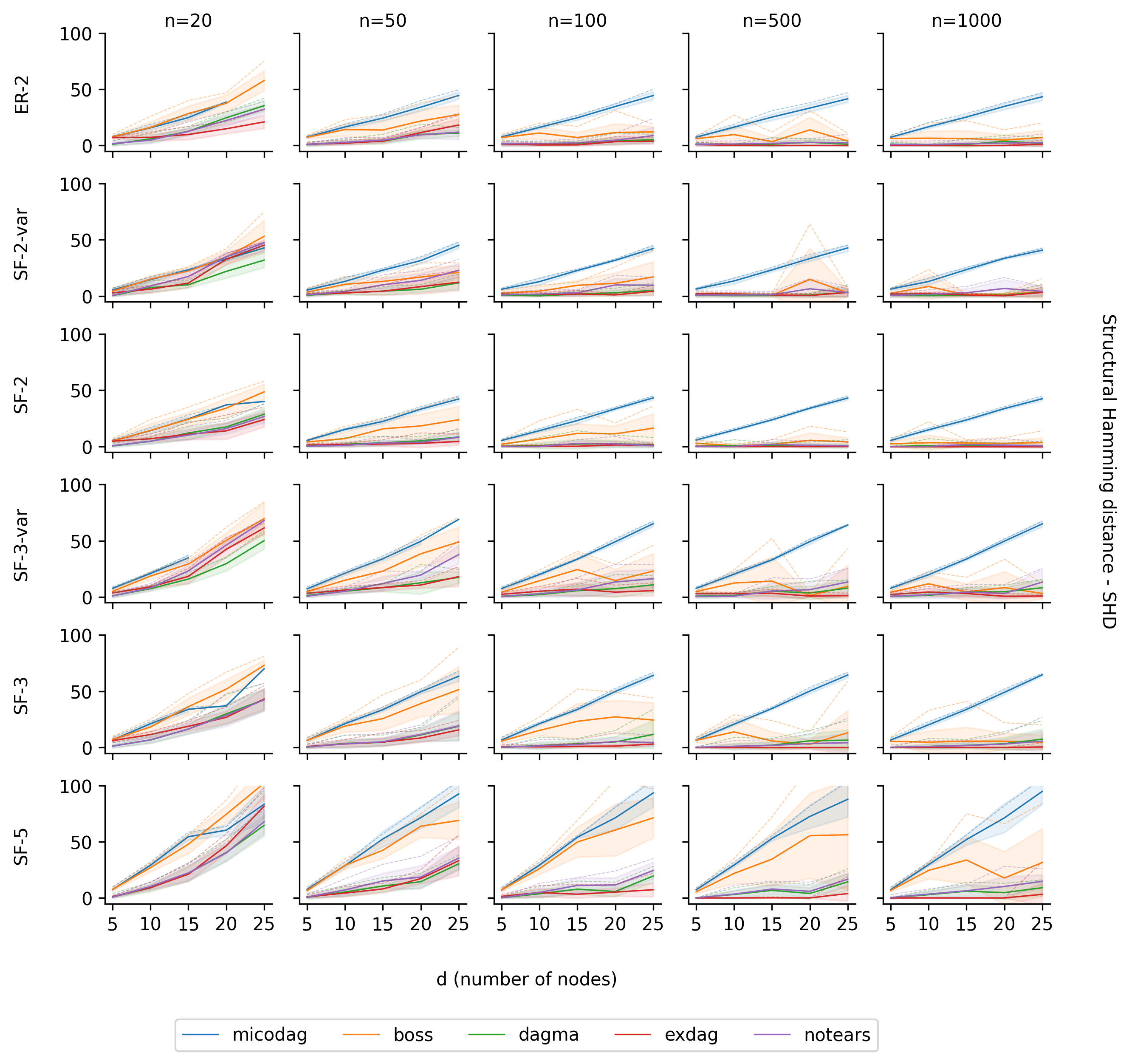}
    \caption{Structural Hamming distance for test cases using ER and SF random ensembles. Mean, standard deviation and minimum over 10 runs is depicted. Standard deviation is depicted as blurred area and the maximum as a dashed line.}
    \label{fig:shd}
\end{figure*}

Notice that once a new mixed-integer feasible solution candidate is found, it is easy to detect cycles therein using the depth-first search (DFS). If a cycle is found, we add the corresponding lazy constraint \eqref{eq_constraints_mi}.
 The DFS algorithm has a worst-case quadratic runtime in the number of vertices of the graph, in contrast to algorithms separating related inequalities from a continuous-valued relaxation \citep{BORNDORFER2020100552,cook2011traveling}, such as the quadratic program in our case. 
 In particular, we have tested three methods of adding lazy constraints: 
\begin{enumerate}
\item Adding only the lazy constraint for the first cycle found. 
\item Adding only the lazy constraint for the shortest cycle found. 
\item Adding multiple lazy constrains for all cycles found in the current integer-feasible solution candidate.
\end{enumerate}
We use Variant 3 throughout our numerical experiments, despite the received wisdom of \citep[Chapter 8.9]{achterberg2007} suggesting that %
one needs to add only a subset of cuts and utilize a carefully crafted selection criterion to identify ``good'' cuts.

\section{Numerical Results}\label{sec_numerics}

In the first comparison, detailed in Section \ref{sec_gaussian_big_comparison}, ExDAG is compared with the state-of-the-art solvers NOTEARS, DAGMA, BOSS, and MICODAG \citep{and24,zheng2018dags,wax24, xu2024integerprogramminglearningdirected}. The experiments are performed under the assumption of Gaussian noise for different generation methods and average edge degrees. Second, it is shown how the convergence to the global minimum allows us to further improve the results of the experiments by granting more computational time. Lastly, we show a simple application to a dataset, which was featured at NeurIPS 2023 competition.

\subsection{Setup Common to all the Benchmarking Experiments and Comparison Metrics}\label{sec_common_setup_for_bench}

\begin{figure*}[h]
    \centering
    \includegraphics[width=0.9\linewidth]{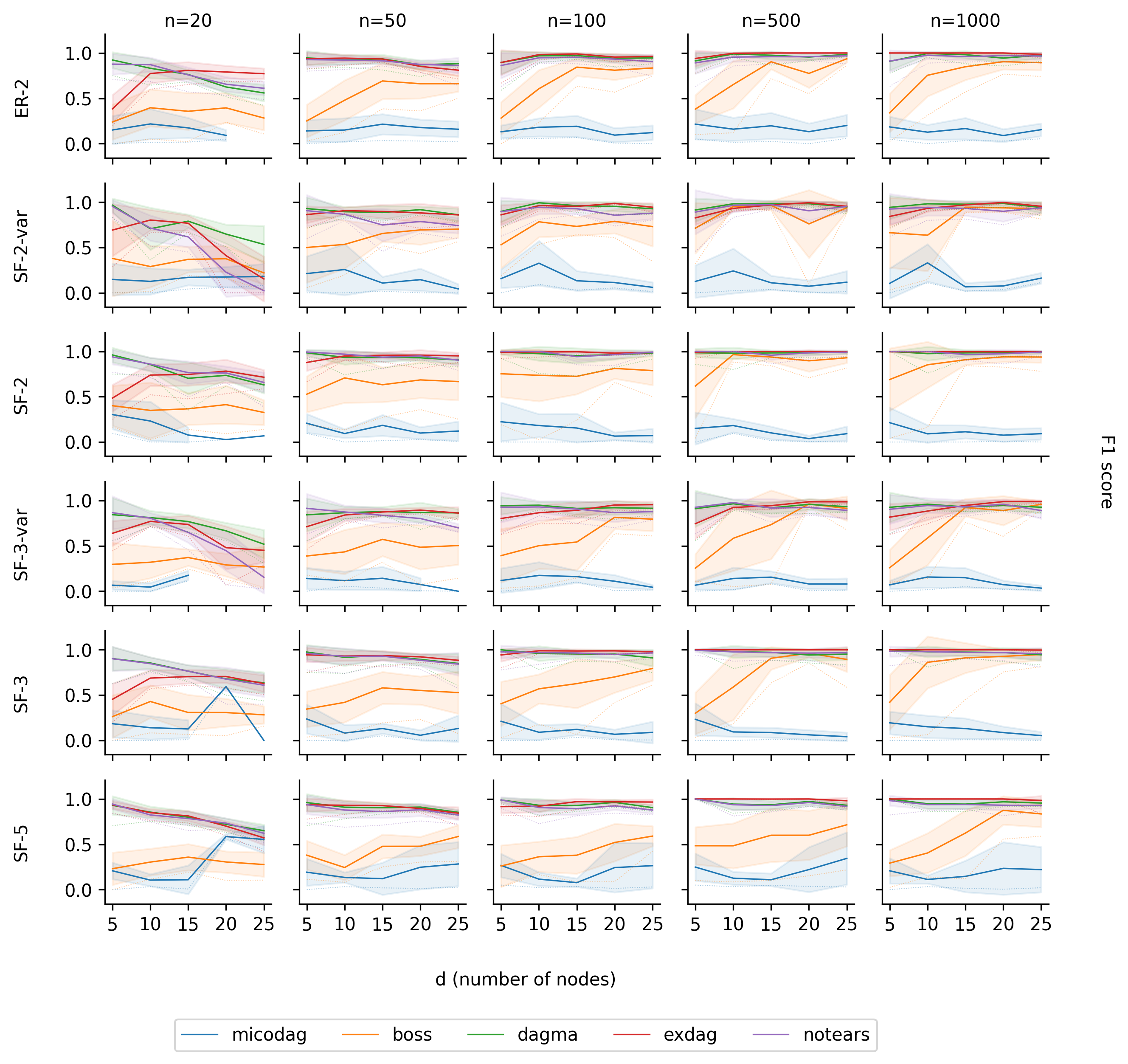}
    \caption{$F_1$ score for test cases using ER and SF random ensembles. Mean, standard deviation and minimum over 10 runs is depicted. Standard deviation is depicted as blurred area and the minimum as a dashed line.}
    \label{fig:f1score}
\end{figure*}

We have implemented a branch-and-bound-and-cut algorithm utilizing Gurobi Optimizer 11, which has been configured to use the simplex algorithm and to use lazy constraints callbacks. These parameter settings are important for three reasons. The simplex algorithm produces corner points of the polyhedra given by \eqref{eq:w} and any of the lazy constraints already added. The corner points of the continuous-valued relaxation can be cut off by constraints \eqref{eq_constraints_mi}, in contrast to points in the interior of the optimal face, which would be obtained by a barrier solver \citep{gondzio2012interior}.  
Second, when Gurobi expects lazy constraints, it avoids pruning the branch-and-bound-and-cut tree prematurely, which would have impacted the global convergence properties \citep{SAHINIDIS1991481} otherwise.
Third, lazy constraints are added directly to LP relaxation, without going through the cut filtering process \citep[Chapter 8.9]{achterberg2007}.
The Python source code is provided in the Supplementary Material and will be open-sourced upon acceptance.  
In the following, we refer to the implementation as ExDAG.

In the experiments with generated data, we generated the data in the following way. First, we generate a ground truth directed acyclic graph using two well-known ensembles of random graphs. We used either the scale-free Barabási–Albert (SF) \cite{barabasi1999emergence}, or the Erd\H{o}s-R\'eny (ER) model \cite{erdHos1960evolution}. Then, the graph weights are sampled uniformly from the set \([-2.0, -0.5] \cup [0.5, 2.0]\). Then, the data samples are generated using the structural equation~\ref{eq:sem}. For each variable, the noise is drawn from a Gaussian distribution with a mean of 0 and a variance of either 1 or a value sampled randomly uniformly from the interval $(0.4, 1.2)$.

To evaluate an algorithm, we run the algorithm on the given dataset to estimate the DAG $W$. 
The structure of the output adjacency matrix $W$ often captures spurious relationships, which can result in an edge with negligible weight in the solution $W$ \citep{NIPS2009_92fb0c6d, pmlr-v48-wange16}. This effect is mitigated by setting a threshold parameter near zero $\delta > 0$, and then setting all elements of $W$ smaller than $\delta$ to $0$. 

Finally, we evaluate the solution $W$ using the following metrics.
Let $A$ and $B$ be two adjacency matrices. Then the structural Hamming distance (SHD) is defined as
\begin{equation}
\text{SHD}\left(A,B\right)=\sum_{i,j=1}^{d}r_{ij}\left(A,B\right),
\end{equation}
where
\begin{equation}
r_{ij}\left(A,B\right)=\begin{cases}
0 & \text{if }A_{ij}\neq0 \neq B_{ij}\text{ or }A_{ij}=0=B_{ij}\\
\frac{1}{2}&\text{if }A_{ij}\neq0\text{ and }B_{ji}\neq0
\\1&\text{otherwise}.
\end{cases}
\end{equation}
SHD is used as a score that quantifies the similarity of two DAGs in terms of edge placement and is commonly used to assess the quality of solutions \citep{zheng2018dags, cussens2012bayesian, Pamfil2020DYNOTEARSSL}. For the solution $W$, we calculate $\text{SHD}(W,W_\text{gt})$, structural Hamming distance between the ground truth $W_\text{gt}$ and the solution $W$, 

In addition to SHD, the F1 score is calculated to evaluate the solution, it is defined as
\begin{equation}\label{eq_F1}
F_{1}=\frac{2}{\text{precision}^{-1}+\text{recall}^{-1}},
\end{equation}
where
\begin{equation}\label{eq_precision}
\text{precision}=\frac{\text{true positive}}{\text{true positive}+\text{false positive}},
\end{equation}
\begin{equation}\label{eq_recall}
\text{recall}=\frac{\text{true positive}}{\text{true positive}+\text{false negative}}. 
\end{equation}

Each algorithm is evaluated in 10 different random instances and the mean, standard deviation, and the maximum for SHD, and minimum for $F_1$ score are reported.

All the experiments in the following section were performed on a computing cluster with AMD EPYC 7543 cpus. We set a limit of 32~GB RAM and two cores per task.

\subsection{Comparison of Identification Methods Under Gaussian Noise}\label{sec_gaussian_big_comparison}

The F1 score and SHD of the graphs identified for the generation methods ER2, SF2, SF3, and SF5 with Gaussian noise are depicted in Figures \ref{fig:shd} and \ref{fig:f1score}. The number after EF and SF denotes the number of edges and vertices ratio. If the name of the problem has "var" suffix, then the variance of the noise of each variable was sampled uniformly from the interval $(0.4,1.2)$. Note that the figure is missing some data points. That means that the method did not finish in the time limit of two hours or crashed.
Summarizing the results, one can see that ExDAG is performing the best most of the time, followed closely by Dagma, and NoTEARS. MICODAG performs well only when the number of samples is low. BOSS performs well only when the number of samples is high and the edges-vertices ratio is low.

Note that the performance of ExDag is influenced by the choice of the regularization coefficient $\lambda$. The coefficient usually have to be chosen per sample count $n$ since the number of samples affects the ratio of the loss minimizing and regularizing part of the cost function \eqref{eq_cost_function_with_p}. The right coefficient can be found by using the following considerations. The first step was a selection of a wide enough set of hyper-parameters $\lambda \in \left\{ 4,3,2,1,0.1,0.05,0.01\right\}$. Then, a multitude of experiments with a short time limit (15 minutes) were performed in which the rate of the decrease of the dual MIP GAP (see definition below) was observed. The ones with the fastest rates of decrease were chosen for the comparison in which the computation time limit was set to 2 hours. The aforementioned strategy may be replicated for cases in which the ground truth is not known.

The MIP GAP is defined as follows
\begin{equation}
\text{MIP GAP}=\frac{\left|J\left(x^{*}\right)-J_{\text{dual}}\left(y^{*}\right)\right|}{\left|J\left(x^{*}\right)\right|},    
\end{equation}
where $x^{*}$ and $y^{*}$ are the current solutions of the primal and dual problems respectively, and $J(x^{*})$ and $J_{\text{dual}}(y^{*})$ are the corresponding utility values.

\subsection{Finding the Optimal Running Time}
\begin{figure}
    \centering
    \includegraphics[width=0.24\linewidth]{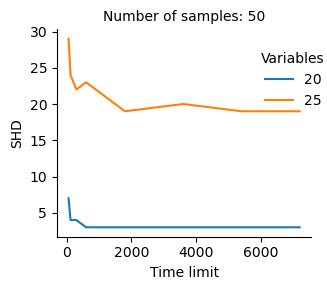}
    \includegraphics[width=0.24\linewidth]{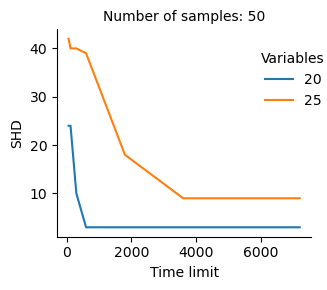}
    \includegraphics[width=0.24\linewidth]{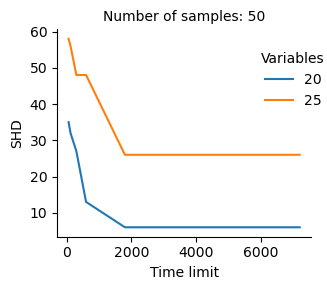}
    \includegraphics[width=0.24\linewidth]{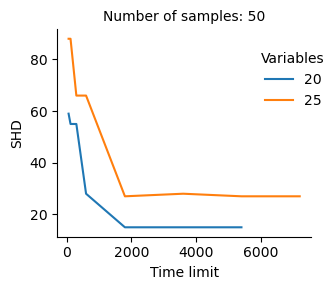}

    \caption{Convergence progress for the generation methods SF2, ER2, SF3, SF5}\label{fig:global_convergence}
\end{figure}

The global convergence of the solver, guaranteed by the use of the branch-and-bound-and-cut (B\&C) algorithm \citep{mitchell2002branch}, needs to also be studied numerically. The progress towards global optimality is monotone, but may stall at times. The information about this progress suggests optimal running times for different problems. In Figure~\ref{fig:global_convergence}, we show that the first plateau was reached relatively quickly in one of the experiments of the previous section. Thus, a time limit of 2000 seconds would have sufficed to arrive at the SHD reported.

\subsection{Identifying DAGs from Datasets with Real Interpretation using ExDAG}
To test the capabilities of ExDAG further, we use it to learn a DAG from \texttt{alarm.csv}, the only publicly available dataset from a competition held at NeurIPS 2023\footnote{Cf. \url{https://codalab.lisn.upsaclay.fr/forums/13855/2071/}}. ExDAG obtains a best SHD of 55 with $\lambda = 0.5$, which improves upon NOTEARS substantially, where NOTEARS identifies DAG with the best SHD score of 65 over 100 different seeds. 
Notice that the identifiability in this case is not well understood. Indeed, depending on the spectral properties of the system, a sufficient number of samples may or may not be sufficient \citep{simchowitz2018learning}
for identifiability. 
Furthermore, the maximum likelihood estimator is not well understood, when the data points are from $\left\{ 0,1\right\} $ and the range of the noise is also $\left\{ 0,1\right\}$. %

\section{Conclusion}
A novel, cycle-based formulation for identifying static Bayesian networks based on the structural vector autoregressive model was proposed. This formulation leads to a mixed-integer quadratic program (MIQP), whose solution is the maximum likelihood estimator for a variety of noise distributions. 
The cycle-based formulation of the problem allows us to add valid cycle-exclusion constraints only upon violation. 
Although the separation of cycle-based inequalities from continuous-valued relaxations is NP-Hard in some settings \citep{BORNDORFER2020100552}, and only heuristics are known \citep{cook2011traveling,vo2023improving} in other settings, our approach is inspired by generalized Benders decomposition for MIQP \citep{geoffrion1972generalized} and the generation of subtour elimination constraints from integer solutions \citep{aguayo2018solving} for the travelling salesman problem \citep{cook2011traveling}, 
which makes it possible to have a separation method with quadratic complexity in the number of vertices of the DAG, i.e., random variables. 

We demonstrated that our method allows for a robust and near-exact reconstruction of DAGs up to 25 vertices, when a sufficient number of samples are available, which surpasses the state of the art (see Section \ref{sec_gaussian_big_comparison}).

Another advantage compared to the state-of-the-art is that our method guarantees global convergenge, which would useful in problems that lead to a nonconvex cost function.

As an important step for further work, one could consider decompositions from structural graph theory 
\citep{hlinveny2008width}, utilized similarly to their use in \emph{a priori} enumeration of cycles in 
\citep{studeny2021dual}.
Similarly, one could consider certain pre-processing of the instances utilizing conditional independence.

\section*{Acknowledgments}
The authors acknowledge the support of National Recovery Plan funded project MPO 60273/24/21300/21000 CEDMO 2.0 NPO.

\bibliographystyle{plain}
\bibliography{ref}

\end{document}